\documentclass[letterpaper, 10 pt, conference, twocolumn]{ieeeconf} 

\usepackage{times}
\usepackage{color, colortbl, xcolor}

\newcommand{\tvar}{t}
\newcommand{\tdummy}{\tau}
\newcommand{\R}{\mathbb{R}}
\newcommand{\ctrl}{u}
\newcommand{\dstb}{d}
\newcommand{\cfunc}{u(\cdot)}

\newcommand{\cset}{\mathcal{U}}

\newcommand{\dset}{\mathcal{D}}

\newcommand{\state}{x}
\newcommand{\traj}{\xi} 

\newcommand{\dyn}{f} 
\newcommand{\targetfunc}{l}
\newcommand{\targetset}{\mathcal{L}}

\newcommand{\costfunctional}{J}

\newcommand{\vfunc}{V}
\newcommand{\vset}{\mathcal{V}}

\newcommand{\trajstandard}{\traj_{\state,\tvar}^{\ctrl}}
\newcommand{\trajstandardwithd}{\traj_{\state,\tvar}^{\ctrl,\dstb}}

\newcommand{\prestate}{\state}
\newcommand{\poststate}{\state^{+}}
\newcommand{\switchsurface}{S}
\newcommand{\resetmap}{\Delta}
\newcommand{\limitcycle}{\state^*}
\newcommand{\period}{T}
\newcommand{\RoA}{\Omega}

\newcommand{\horizon}{T}

\newtheorem{remark}{Remark}

\newtheorem{theorem}{Theorem}

\newcommand{\baselinectrl}{\pi_0}
\usepackage{multicol}
\usepackage[bookmarks=true]{hyperref}

\usepackage{amsfonts}
\usepackage{algorithmic}
\usepackage[ruled,vlined,titlenotnumbered]{algorithm2e} 
\usepackage{amsmath}
\usepackage{dsfont}
\usepackage{subfigure}
\usepackage{multicol}
\usepackage[pdftex]{graphicx}   
\usepackage{graphicx}
\usepackage{bbm}
\usepackage{mathtools}
\usepackage{dsfont}
\usepackage{accents}
\usepackage{wrapfig}
\usepackage{siunitx}

\usepackage{url}

\graphicspath{{./figs/}}    
\DeclareGraphicsExtensions{.pdf,.png}  

\usepackage[textfont=md,font=footnotesize]{caption}

\usepackage[left=55pt, right=55pt, top=55pt, bottom=58pt]{geometry} 

\IEEEoverridecommandlockouts
\usepackage{balance}

\usepackage[%
    style=numeric-comp,
    sorting=none,
    backend=biber,
    sortcites=true,
    doi=false,
    firstinits=true,
    hyperref,
    isbn=false,
    eprint=false,
    maxcitenames=3, 
    minbibnames=3, 
    maxbibnames=4, 
    block=none]
    {biblatex}
    
    \renewbibmacro{in:}{}
    \AtEveryBibitem{%
  	\clearlist{language}%
  	\clearfield{pages}%
	}
\bibliography{references,litreview}
\usepackage{lipsum}

\definecolor{planning_color}{RGB}{69, 174, 254}    
\definecolor{prediction_color}{RGB}{255, 116, 190}
    
\newcommand{\example}[1]%
{
\textbf{Running example:}
\textit{#1}
}

\newcommand{\jcnote}[1]{\textcolor{black}{#1}}

\title{Computation of Regions of Attraction for Hybrid Limit Cycles Using Reachability: An Application to Walking Robots}
\author{Jason J. Choi, Ayush Agrawal, Koushil Sreenath, Claire J. Tomlin, and Somil Bansal
\thanks{
\jcnote{This research is supported in part by NSF Grants CMMI-1931853, CMMI-1944722, an ONR Basic Research Challenge Program on Multibody Systems, NASA ULI program (through Stanford) on Safe Aviation Autonomy, and the DARPA Assured Autonomy Program. The work of Jason Choi received the support of a fellowship from Kwanjeong Educational Foundation.} Jason Choi, Ayush Agrawal, Koushil Sreenath, and Claire Tomlin are with the University of California, Berkeley, and Somil Bansal is with the University of Southern California, Los Angeles. Contact info: \{jason.choi, ayush.agrawal,
koushils, tomlin\}@berkeley.edu, somilban@usc.edu. }
\thanks{Project website: \url{https://hybridreachability.github.io/website-hybrid-roa/}}
}

\begin{document}

\maketitle

\renewcommand{\baselinestretch}{1.00}

\begin{abstract}
Contact-rich robotic systems, such as legged robots and manipulators, are often represented as hybrid systems. 
However, the stability analysis and region-of-attraction computation for these systems are often challenging 
because of the discontinuous state changes upon contact (also referred to as \textit{state resets}). 
In this work, we cast the computation of region-of-attraction as a Hamilton-Jacobi (HJ) reachability problem.
This enables us to leverage HJ reachability tools that are compatible with general nonlinear system dynamics, and can formally deal with state and input constraints as well as bounded disturbances. Our main contribution is the generalization of HJ reachability framework to account for the discontinuous state changes originating from state resets, which has remained a challenge until now.
We apply our approach for computing region-of-attractions for several underactuated walking robots and demonstrate that the proposed approach can (a) recover a bigger region-of-attraction than state-of-the-art approaches, 
(b) handle state resets, nonlinear dynamics, external disturbances, and input constraints, and (c) also provides a stabilizing controller for the system that can leverage the state resets for enhancing system stability.
\end{abstract}

\IEEEpeerreviewmaketitle

\section{Introduction}
Practical legged locomotion is one of the fundamental problems in robotics.
Legged robots can exhibit efficient, versatile, and highly dynamic gaits that allow them to operate in highly unstructured environments.
Mathematically, legged robots are often represented as hybrid dynamical systems.
Hybrid dynamical systems may exhibit both continuous and discrete dynamics, wherein swinging of a leg has continuous dynamics, whereas contacts with ground are well-modeled as discrete events \cite{westervelt2018feedback}.
Under hybrid models, walking gaits are represented as periodic \textit{hybrid limit cycles}, i.e., a limit cycle involving discontinuous jumps in the system state when the leg impacts the ground.

To obtain a stable walking behavior, it is often desirable to compute the \textit{region-of-attraction (RoA)} for a walking gait -- the robot configurations from which it can eventually converge to the corresponding hybrid limit cycle and follow the gait.
Other than analyzing the stability of a gait, RoA estimation can also be useful for designing different candidate control laws for walking and switching between different gaits \cite{motahar2016composing}.
Despite significant progress over the last few years, stability analysis and RoA computation for legged robots is still quite challenging because the dynamics are highly nonlinear and intrinsically hybrid, requiring controlling through contact. 

In this work, we cast the RoA computation as a Hamilton-Jacobi (HJ) reachability problem. 
In reachability analysis, one computes the \textit{Backward Reachable Tube (BRT)} of a dynamical system -- the set of states such that the trajectories that start from this set will eventually reach some given target set despite the worst case disturbance \cite{bansal2017hamilton}.
If we use the limit cycle as the target set, the corresponding BRT is the RoA of the limit cycle. The advantage of casting RoA computation as a reachability problem is the availability of numerical tools that can compute BRT for a variety of nonlinear systems in the presence of disturbances and control bounds. 
Even though computationally intensive compared to other existing methods that can solve for the underapproximations of RoAs, the reachability-based methods recover the maximal RoA without imposing any restriction on its shape.

However, one problem that still remains is that, so far, we lack a general framework for the reachability analysis of dynamical systems that involve discontinuous state changes (also called \textit{state resets)}.
To overcome this challenge, we extend the reachability framework to dynamics with discontinuous state resets. 
Our key insight is that since HJ reachability analysis ultimately corresponds to solving a dynamic programming problem, we can reason about the effect of state resets on the BRT by applying the Bellman principle of optimality at the switching surface.
We demonstrate that this corresponds to a value remapping between post and pre reset states during the BRT computation. 
Moreover, the existing numerical algorithms to compute BRTs can easily be extended to account for this remapping without incurring any additional computation cost. Thus, the obtained value function, the corresponding BRT, as well as the stabilizing controller implicitly reason about the effect of state resets on the stability of the system. 

We apply the proposed approach on various ``dynamic walking'' systems and demonstrate that it can obtain a bigger RoA than most state-of-the-art approaches, while accounting for input constraints, state resets, and external disturbances at the same time. 
Finally, the proposed approach also provides a stabilizing controller for the system that can leverage state resets for a faster stabilization.  

\section{Related Work}
\emph{Poincar\'e Map-based methods. } Stability analysis of limit cycles can be reduced to the problem of analyzing stability of the fixed point of the corresponding Poincar\'e map. 
In particular, if the fixed point of the Poincar\'e map is stable, then there exists a stable limit cycle for the original system that passes through this fixed point \cite{guckenheimer1988structurally,grizzle2006remarks}. 
However, finding an analytical expression for the Poincar\'e map is challenging; hence, a common approach is to compute a numerical approximation of its linearization \cite{westervelt2018feedback,morris2005restricted}.
The local stability and RoA of the limit cycle are then computed using this discrete-time linear system \cite{motahar2016composing}.
This approach has also been used to analyze stability in the presence of external disturbances \cite{veer2019input}.

\emph{ Lyapunov-based methods.} One of the limitations of  Poincar\'e analysis is that the verifiable RoA is restricted to the set of states on the chosen Poincar\'e section.
The works in \cite{manchester2011regions,manchester2011transverse,freidovich2008stability} use linearized transverse dynamics to estimate the RoA by constructing a (time-varying) Lyapunov function via Sum-of-Squares (SOS) programs.
As noted in \cite{manchester2011regions}, a limitation of this approach is that the size of the computed RoA depends on the choice of the transverse coordinates. The work in \cite{posa2013lyapunov} also proposes a Lyapunov based approach to compute RoA for nonlinear hybrid systems with state resets using the \emph{complementarity} formulation of contact. 

In several of the aforementioned methods, Lyapunov functions are computed using SOS programming.
The key advantage of SOS-based methods is that they can leverage convex optimization to compute fast underapproximations of the RoA, leading to significant computation gains over reachability-based methods.
However, as a tradeoff, the verified RoA can be a significant underapproximation of the true RoA.
First, the SOS methods restrict the Lyapunov functions to the class of (piecewise) SOS polynomials.
Second, although one can typically design a convex SOS program when the feedback controller is fixed, solving for the controller and the Lyapunov function together to maximize the verifiable RoA results in a nonconvex program which induces another source of underapproximation when we obtain a local optimum.
These limitations are exacerbated further in the presence of state resets as the Lyapunov condition at the switching surface leads to a nonconvex program in general \cite{manchester2011regions}. Ultimately, this results in recovering a smaller RoA.
In contrast, the proposed approach does not impose any specific structure on the value function and hence can recover the maximal RoA (subjected to numerical errors), even in the presence of state resets.

\emph{Stabilizing controllers for hybrid systems with state resets.} 
The \emph{Hybrid Zero Dynamics} (HZD) framework \cite{grizzle2001asymptotically,westervelt2003hybrid,westervelt2018feedback} was one of the earliest works on creating stabilizable hybrid limit cycles and stabilizing controllers for walking robots with discrete impacts. The method of Poincar\'e can also be incorporated into the HZD to develop \emph{event-based} controllers to stabilize or switch between hybrid limit cycles \cite{hamed2016exponentially,shih2012stable,agrawal2017discrete}. Building on the HZD framework, \cite{ames2014rapidly} proposes a \emph{rapidly exponentially stabilizing} control Lyapunov function based quadratic program (CLF-QP) to exponentially stabilize hybrid limit cycles. There is also a rich literature on contact implicit trajectory optimization \cite{posa2014direct,li2020hybrid} and model predictive control \cite{cleac2021linear,marcucci2017approximate} that can simultaneously handle multiple contact phases. 
Some of these methods have also leveraged value remapping to account for the state resets.
In this work, we extend this principle to reachability analysis and focus on the RoA computation, as opposed to trajectory optimization.
\section{Problem Setup} \label{sec:problem_setup}

We consider the following class of hybrid systems:
\begin{equation}
\label{eq:dyn}
\begin{aligned}
\dot{\state} &= \dyn(\state, \ctrl, \dstb), & & \prestate \notin \switchsurface \\
\poststate &=\resetmap\left(\prestate^{-}\right), & & \prestate^{-} \in \switchsurface,
\end{aligned}
\end{equation}
where $\state \in \R^n$ is the state, $\ctrl \in \cset$ is the control input, and $\dstb \in \dset$ is the disturbance.
$\switchsurface$ indicates the switching surface where the reset map $\resetmap$ is applied; under the reset event at time $t$, the state $\prestate^{-} :=\lim_{\tau \nearrow t}x(\tau)$ instantaneously shifts to $\poststate$. 
Therefore, the resulting trajectory of this hybrid system is composed of a set of continuous trajectories along the vector flow $\dyn$, and discontinuous jumps between them whenever the state hits the switching surface $\switchsurface$.

\textbf{\textit{Hybrid limit cycle:}} Let $\limitcycle(\cdot)$ represents a non-trivial $\period$-periodic solution of the system that undergoes $N$ impacts at times $\{t_1, t_2, \ldots, t_N\} + k\period$ for non-negative integer $k$.
We call $\limitcycle(\cdot)$ a hybrid limit cycle of the system.

We assume that this limit cycle is given along with a baseline control law $\baselinectrl(\cdot):\R^n\rightarrow \cset$ that makes the limit cycle forward invariant and stable in the small neighborhood around it. 
There are many existing works that study approaches for designing such stable hybrid limit cycles \cite{westervelt2003hybrid,grizzle2001asymptotically} as well as the locally-stabilizing control laws \cite{ames2014rapidly,westervelt2018feedback}. 

The overall objective is twofold. First, we seek to compute a region of state space $\RoA(\limitcycle) \subseteq \R^n$ around $\limitcycle(\cdot)$, from which there exists an admissible control that can asymptotically stabilize the system to $\limitcycle(\cdot)$.
We call $\RoA(\limitcycle)$ the (asymptotically) stabilizable region or RoA for $\limitcycle(\cdot)$.
Second, we also want to design a corresponding control law that stabilizes the system trajectory to $\limitcycle(\cdot)$.

\begin{remark}
Note that depending on the baseline controller, the RoA under $\pi_0(\cdot)$ can be very small.
Our goal in this work is to find a stabilizing control law that is beyond the capability of the baseline controller and provides us with the \textit{maximal} RoA $\RoA(\limitcycle)$. 
\end{remark}

\textit{\textbf{Running example (Teleporting Dubins Car).}} 

\begin{figure}\centering
\includegraphics[width=\columnwidth]{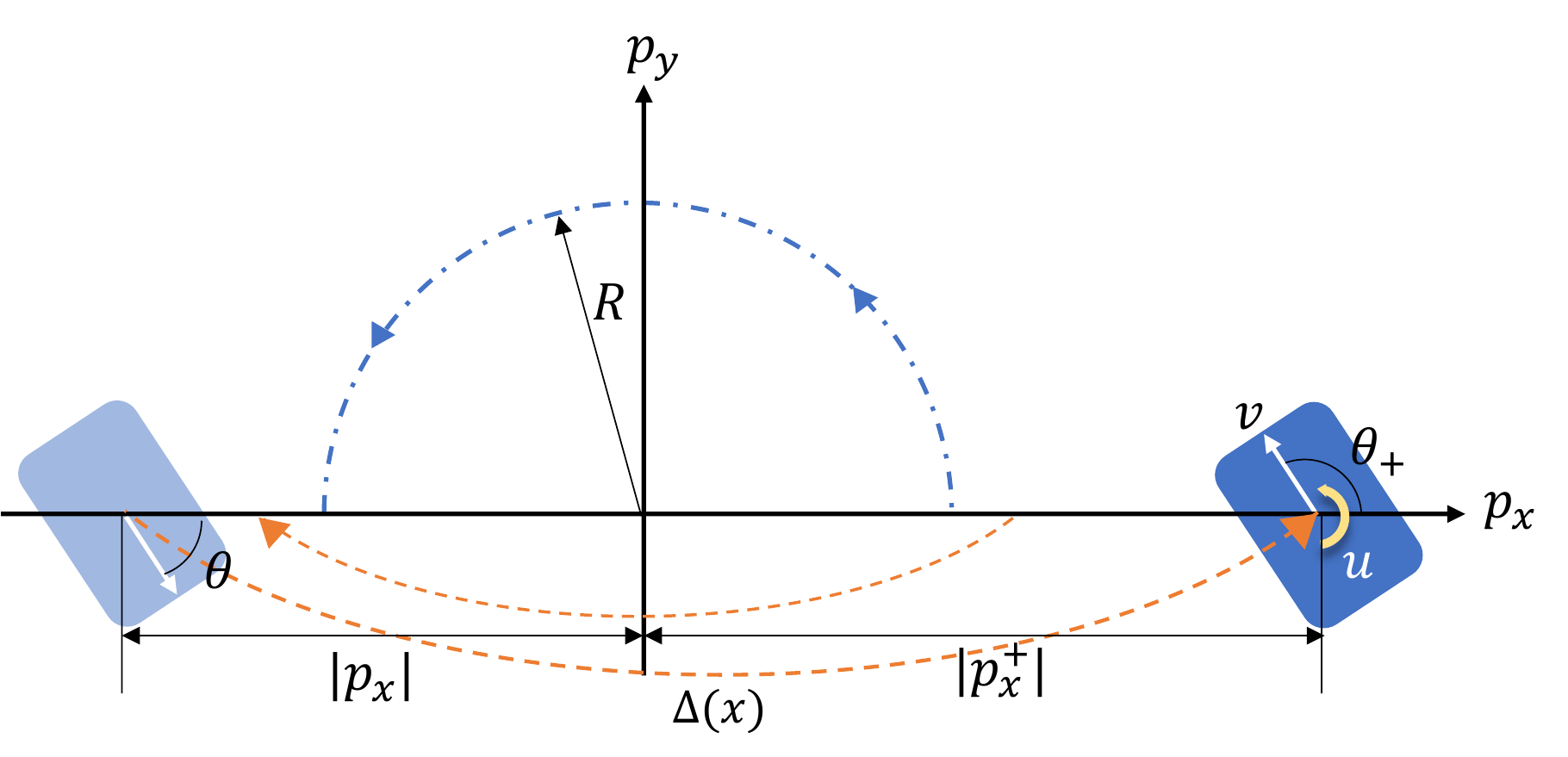}
\vspace{-1em}
\caption{Teleporting Dubins car system. The car ``teleports'' to the other side of the x-axis whenever it hits the x-axis from a positive y-direction.}
\label{fig:dubins_config}
\vspace{-1.5em}
\end{figure}

We now introduce a simple hybrid system example that we will use throughout the paper to illustrate our method. We introduce state resets to a well-known Dubins car system which makes the car ``teleport'' (Fig. \ref{fig:dubins_config}). It is chosen because the key principles of our method are well exposed in its design and results; more practical examples will be introduced in Sec VI. The dynamics are given by
\begin{equation} \label{eqn:dubins_dyn}
\dot{p}_x = v \cos \theta, \quad \dot{p}_y = v \sin \theta, \quad \dot{\theta} = u,
\end{equation}
where $\state = [p_x, p_y, \theta]$ is the state of the system consisting of car's position $(p_x, p_y)$ and its heading $\theta$.
$v$ is the constant velocity of the car, and $u\in[-\bar\omega, \bar\omega]$ is the control input, which is the angular speed. Next, we introduce a switching surface and the reset map as the following:
\begin{align}
    \switchsurface & = \{\state \; | \; p_y = 0, \sin \theta < 0 \} \\
    \quad \poststate & = \resetmap(\prestate^{-}) = \left[
- R \cdot (|p_x|/R)^{\alpha} \cdot \text{sign}(p_x),~~ p_y,~~ \pi + \theta\right] \nonumber
\end{align}

Intuitively, the car undergoes a discrete jump whenever it hits the $x$-axis from a positive $y$-direction.
Upon the reset, the car's $x$-position and heading change discontinuously. 
The parameter $\alpha > 0$ in the reset map dictates the nonlinearity of the reset map. 
For instance, $\alpha = 1$ makes $p_x^+ = -p_x$, i.e., the car position is mirrored around the y-axis.
For $\alpha < 1$, the reset map has a contraction property--everytime the state resets, the value of $|p_x|$ gets closer to $R$. On the other hand, value of $\alpha > 1$ induces an expanding reset map.

The desired hybrid limit cycle is a semi-circular trajectory of radius $R$ centered at the origin in the counterclockwise direction. Therefore, the orbit can be expressed as

\small
\vspace{-1em}
\begin{equation*}
    \limitcycle(\cdot) = \left\{ \state ~\middle|~ \|(p_x, p_y)\|\!=\!R, \;\theta\!=\!\frac{\pi}{2} + \tan^{-1}(\frac{p_y}{p_x}), \frac{\pi}{2} \le \theta < \frac{3\pi}{2} \right\}.
\end{equation*}
\normalsize
We are interested in computing the RoA for $\limitcycle(\cdot)$, i.e., the states from which the car can be stabilized to follow the semi-circular trajectory eventually, as well as the corresponding stabilizing controller. A baseline stabilizing controller $\baselinectrl(\cdot)$ is designed by feedback linearization, with the output defined as $y=\|(p_x, p_y)\| - R$.
\section{Background} \label{sec:background}

\subsection{Hamilton-Jacobi Reachability}
\label{subsec:hj_reachability}
We present an overview of Hamilton-Jacobi (HJ) reachability analysis, first for continuous systems without the disturbance: the dynamics of the system is given by $\dot{x} = f(x, u)$. Let $\trajstandard(\tdummy)$ denote the state achieved at time $\tdummy$ by starting at initial state $\state$ and initial time $\tvar$, and applying input functions $\cfunc$ over $[\tvar,\tau]$. 
Given a target set $\targetset \subset \R^n$, we are interested in computing the Backward Reachable Tube (BRT) of $\targetset$:

\textit{\textbf{Backward Reachable Tube (BRT)}} is the set of initial states from which the agent acting optimally will \textit{eventually} reach the target set $\targetset$ within the time horizon $[\tvar, \horizon]$:
\vspace{-0.25em}
\begin{equation}
\label{eqn:BRT}
\small{
\vset(\tvar) = \{\state \; | \; \exists \ctrl(\cdot):[\tvar, \horizon]\rightarrow\cset, \exists \tdummy \in [\tvar, \horizon], \trajstandard(\tdummy) \in \targetset\}}.
\vspace{-0.25em}
\end{equation}

In HJ reachability, the computation of BRT is formulated as an optimal control problem, which can be solved by using the principle of dynamic programming. 

First, a Lipschitz continuous target function $\targetfunc(\state)$ is defined whose zero sub-level set is the target set $\targetset$, i.e. $\targetset = \{\state : \targetfunc(\state) \leq 0\}$.  Typically,  $\targetfunc(\state)$ is defined as a signed distance function to $\targetset$.
The BRT seeks to find all states that could enter $\targetset$ at any point in the time horizon.
This is computed by finding the minimum distance to $\targetset$ over time:

\small
\vspace{-0.75em}
\begin{equation}
    \label{eq:costfunctional}
    \costfunctional(\state,\tvar,\cfunc) = \min_{\tdummy \in [\tvar, \horizon]} \targetfunc(\trajstandard(\tdummy)).
    \vspace{-0.25em}
\end{equation}
\normalsize
Our goal is to capture this minimum distance for \textit{optimal trajectories} of the system.  
Thus, we compute the optimal control that minimizes this distance (and eventually drives the system into the target set when $\targetfunc(\cdot)\le0$). The value function corresponding to this optimal control problem is:
\vspace{-0.25em}
\begin{equation}
\small{
    \label{eq:valuefunc}
    \vfunc(\state,\tvar) = \inf_{\cfunc} \Big\{\costfunctional\Big(\state,\tvar,\cfunc\Big)\Big\}.}
    \vspace{-0.25em}
\end{equation}
\normalsize
The value function in \eqref{eq:valuefunc} can be computed using dynamic programming, which results in the following final value Hamilton-Jacobi Isaacs Variational Inequality (HJI-VI):

\small
\vspace{-0.5em}
\begin{equation}
\begin{aligned}
    \label{eq:HJIVI}
    \min\Big\{D_\tvar \vfunc(\state,\tvar)+ H(\state,\tvar), \targetfunc(\state)-\vfunc(\state,\tvar)\Big\} = 0,
    \end{aligned}
    \vspace{-0.25em}
\end{equation}
\normalsize
with the terminal value function $\vfunc(\state,\horizon) = \targetfunc(\state)$. 
$D_\tvar$ and $\nabla$ represent the time and spatial gradients of the value function. 
$H$ is the Hamiltonian that encodes the role of dynamics and the optimal control inputs:

\small
\vspace{-0.5em}
\begin{equation}
    \label{eq:ham}
    \begin{aligned}
    H(\state,\tvar) = \min_\ctrl & \langle \nabla \vfunc(\state,\tvar), \dyn(\state,\ctrl)\rangle.
        \end{aligned}
    \vspace{-0.25em}
\end{equation}
\normalsize
Intuitively, the term $\targetfunc(\state)-\vfunc(\state,\tvar)$ in \eqref{eq:HJIVI} enforces the value function to ``memorize'' the best record (closest instance to the target set) of the optimal trajectories. Once $\vfunc(\state,\tvar)$ is obtained, the BRT is given as the zero sub-level set of the value function
$\vset(\tvar) = \{\state \; | \; \vfunc(\state,\tvar) \leq 0 \}$. The corresponding optimal control for reaching the target set $\targetset$ is derived as

\small
\vspace{-0.25em}
\begin{equation}
    \label{eq:opt_ctrl}
    \begin{aligned}
    u^*(\state,\tvar) = \arg\min_\ctrl  & \langle \nabla \vfunc(\state,\tvar), \dyn(\state,\ctrl)\rangle.
        \end{aligned}
    \vspace{-0.25em}
\end{equation}
\normalsize
Finally, note that every state that is contained in a \textit{finite}-time BRT of $\targetset$, represents an initial condition from which the system can be controlled to $\targetset$ within finite time.

\subsection{Casting RoA computation as a HJ reachability problem}
\label{subsec:roa_hj}

We now explain how HJ reachability can be used to estimate the RoA for a hybrid limit cycle, $\RoA(\limitcycle)$. 
Let $h(x)$ denote distance to the limit cycle $\limitcycle(\cdot)$, i.e., $h(x) := \inf_{y\in \limitcycle(\cdot)} \|x - y\|$.
We define the target function as $\targetfunc(x) := h(x) - \epsilon$ for some $\epsilon > 0$.
Thus, the target set $\targetset$ is an $\epsilon$-tube around the orbit $\limitcycle(\cdot)$.
Note that for a stable hybrid limit cycle, we can pick $\epsilon$ such the baseline controller $\pi_0$ can stabilize to the limit cycle from anywhere inside $\targetset$.

Given the value function, $\vfunc(\state,\tvar)$ and the BRT, $\vset(\tvar)$, corresponding to this target set, we can stabilize to the limit cycle from any state within the BRT using the following control law:
\vspace{-0.5em}
\begin{equation} \label{eqn:overall_stabilizing_controller}
   u(x, t)= 
\begin{cases}
   \arg\min_\ctrl \langle \nabla \vfunc(\state,\tvar), \dyn(\state,\ctrl)\rangle,& \text{if } x \notin \targetset\\
    \pi_0(x),              & \text{otherwise}
\end{cases} 
\end{equation}
Intuitively, we can use the controller provided by HJ reachability to steer the system to $\targetset$ from any state within the BRT; upon reaching $\targetset$, we can switch to the baseline controller. 
Therefore, for every state that can be verified as an element of a finite-time BRT of $\targetset$, we can conclude that it is an element of $\RoA(\limitcycle)$.
In other words, $\vset(\tvar)$ can be used as an estimation of $\RoA(\limitcycle)$.
The corresponding stabilizing controller is given by \eqref{eqn:overall_stabilizing_controller}.

\begin{remark} \label{remark:baseline_policy_need}
Note that if we pick $\epsilon = 0$, only the states that can achieve finite-time convergence to $\limitcycle(\cdot)$ can be verified from HJ reachability analysis. However, this would exclude many states that are not finite-time stabilizable but still asymptotically stabilizable to $\limitcycle(\cdot)$. In fact, even for a state where stronger notion like exponential stabilizability \cite{ames2014rapidly} is achievable, its time-to-reach to the orbit is not necessarily a finite value.
Therefore, by using $\epsilon > 0$ together with the baseline policy $\pi_0$ which makes sure that every state in $\targetset$ is stabilizable to the actual orbit, we can capture the states that are not necessarily finite-time but are still asymptotically stabilizable to $\limitcycle(\cdot)$ and are inside the finite-time BRT of $\targetset$. For more detailed analysis and a constructive way to choose $\epsilon$, please refer to \cite{clf_zubov}. Finally, defining $\targetset$ as the $\epsilon$-neighborhood of $\limitcycle(\cdot)$ is also necessary for numerical methods to compute $\vfunc(\state, \tvar)$ and to tackle the presence of disturbance, which will be explained in the next subsection.
\end{remark}

\textit{\textbf{Running example (Teleporting Dubins Car).}} The target function $\targetfunc(\state)$ for the running example is  $\targetfunc(\state) := \|(p_x, p_y)\| - R - \epsilon$.
The Hamiltonian can be computed analytically in this case (see \cite{Mitchell05}):
\begin{equation}
    \label{eq:ham_dubins}
    H(\state,\tvar) = \frac{\partial \vfunc}{\partial p_x} v \cos \theta + \frac{\partial \vfunc}{\partial p_y} v \sin \theta - \bar{\omega}|\frac{\partial \vfunc}{\partial \theta}|.
\end{equation}

\subsection{Robustness to bounded disturbance}
\label{subsec:disturbance}

For systems with bounded disturbance $d\in\dset$, HJ reachability can again be used to synthesize optimal controllers that are robust to the disturbance. 
The definition of BRT is now the set of initial states for which, \textit{under worst-case disturbances}, the agent acting optimally will eventually reach the target set $\targetset$ within time $[\tvar, \horizon]$:
\vspace{-0.25em}
\begin{equation}
\label{eqn:BRT_d}
\small{
\vset(\tvar) = \{\state: \exists \ctrl(\cdot), \forall \dstb(\cdot):[\tvar, \horizon]\rightarrow \dset, \exists \tdummy \in [\tvar, \horizon], \trajstandardwithd(\tdummy) \in \targetset\}}.
\vspace{-0.25em}
\end{equation}

Computation of the BRT can be formulated as a zero-sum game between the control and the disturbance, and can be solved in a similar way by applying the dynamic programming \cite{Mitchell05}. $V(x, t)$ is governed by the same variational inequality as \eqref{eq:HJIVI}, where the Hamiltonian is as follows:

\small
\vspace{-0.5em}
\begin{equation}
    \label{eq:ham_d}
    \begin{aligned}
    H(\state,\tvar) = \max_\ctrl \min_\dstb & \langle \nabla \vfunc(\state,\tvar), \dyn(\state,\ctrl,\dstb)\rangle.
        \end{aligned}
    \vspace{-0.5em}
\end{equation}
\normalsize
Finally, the optimal control signal that is robust to the disturbance can be synthesized by the following rule:

\small
\vspace{-0.5em}
\begin{equation}
    \label{eq:opt_ctrl_d}
    \begin{aligned}
    u^*(\state,\tvar) = \arg\max_\ctrl \min_\dstb & \langle \nabla \vfunc(\state,\tvar), \dyn(\state,\ctrl,\dstb)\rangle.
        \end{aligned}
    \vspace{-0.5em}
\end{equation}
\normalsize

For the specific problem of stabilization to the hybrid limit cycle, it must be noted that any non-zero disturbance signal can break the forward-invariance of the baseline controller $\pi_0(\cdot)$ on the limit cycle $\limitcycle(\cdot)$. Therefore, the notion of $\limitcycle(\cdot)$ being \textit{robustly asymptotically stable} is defined as an existence of an $\epsilon$-neighborhood of the limit cycle to which the system can be asymptotically stabilized under the worst-case disturbance \cite{freeman1996inverse}. Under a mild assumption that the baseline controller $\baselinectrl(\cdot)$ itself renders the set $\targetset$ locally robustly stable, any elements of the finite-time BRT for $\targetset$ can be verified as robustly asymptotically stabilizable to $\limitcycle(\cdot)$. 
\section{Hamilton-Jacobi Reachability for Dynamics With State Resets} \label{sec:approach}
We now present the main contribution of this paper, which is an extension of the reachability framework above (for continuous dynamics) to account for discontinuous state resets. First, the
following theorem proves that the value function in the presence of state resets can be obtained by solving a constrained version of HJI-VI in \eqref{eq:HJIVI}.
\begin{theorem} \label{theorem1}
$\vfunc(\state,\tvar)$ in the presence of state resets is given by the solution to the following constrained VI:

\small
\vspace{-1.5em}
\begin{equation}
\begin{aligned}
    \label{eq:HJIVI_constrained}
    \min\Big\{D_\tvar \vfunc(\state,\tvar)+ H(\state,\tvar), \targetfunc(\state)-\vfunc(\state,\tvar)\Big\} = 0, & & \prestate \notin \switchsurface, \\
    \vfunc(\state,\tvar) = \vfunc(\resetmap\left(\prestate\right),\tvar), & & \prestate \in \switchsurface,
    \end{aligned}
\end{equation}
\normalsize
with the terminal value function 
\begin{equation}
\begin{aligned}
    \label{eq:HJIVI_constrained_IC}
    \vfunc(\state,\horizon) & = \targetfunc(\state), & & \prestate \notin \switchsurface, \\
    \vfunc(\state,\horizon) & = \targetfunc(\resetmap\left(\prestate\right)), & & \prestate \in \switchsurface.
    \end{aligned}
\end{equation}
\vspace{-1.0em}
\end{theorem}
Intuitively, Theorem \ref{theorem1} states that the value function can be obtained by solving the usual HJI-VI for the states that are not on the switching surface. 
For the states that are on the switching surface, the value is given by that of the corresponding post reset state.
This is not surprising because if the state is on the switching surface, it will instantaneously change to the post reset state.
Consequently, how ``close'' the state is to the target set is completely characterized by how close the post reset state is to the target set. 
The proof of Theorem \ref{theorem1} formalizes this intuition using the Bellman principle of optimality and is presented in Appendix A in the supplementary material.

The interesting aspect about the HJI-VI in \eqref{eq:HJIVI_constrained} is that since it reasons about the state resets, the obtained value function and the stabilizing controller implicitly exploit the reset map to reach the target set as quickly as possible.
We will demonstrate this aspect further in Sec. \ref{sec:dubins_car:results}. 

\textit{\textbf{Numerical Implementation.}} We now present a numerical method to compute the value function based on the result in Theorem \ref{theorem1}. 
The baseline of the numerical algorithm is a standard level set-based method that computes the value function over a discretized computational domain of a compact subset $\mathcal{X} \subset \R^n$ \cite{mitchell2004toolbox}. 
Let $\{p(x)\}_G$ denote the set of values of a function $p(x)$ evaluated over a state space grid $G$.
Since our computation will proceed backward in time, we will numerically solve the variational inequality in \eqref{eq:HJIVI_constrained} for each discrete time step in a finite interval $[0, T]$ using the three-step update rule described in Algorithm \ref{numerical_algo}.
\begin{algorithm}[htb]
\SetAlgoLined
\SetKwInOut{Input}{Input}
\SetKwInOut{Output}{Output}
\Input{Target function $\{l(x)\}_G$, Time horizon $T$}
\Output{Value function $\{V(x,t)\}_G$ for $t = 0, \delta, 2\delta, \cdots, T$}
Initialization: $\{V(x,T)\}_{G \setminus S}= \{l(x)\}_{G \setminus S}$,\\
~~~~~~~~~~~~~~~~~$\{V(x,T)\}_{S}= l(\poststate)$,\\
~~~~~~~~~~~~~~~~~$t = T$\;
 \While{$t > 0$}{
 $\text{For all } x\in G \setminus S,$
\begin{align} \label{ham_in_algo}
V(x, t-\delta) = & V(x, t)\\
& + \min_{u\in \cset} \max_{d\in \dset} D_x V(x, t) \cdot f(x, u) \;\delta; \nonumber
\end{align}
$V(x, t-\delta) = \min\{V(x, t-\delta), l(x)\}$;\\
$\text{For all } x\in S,$
\begin{align}  \label{constraint_in_algo}
V(x, t-\delta)= V(\Delta(x), t-\delta);
\end{align}
\\
$t = t - \delta$\;
 }
 \caption{Value Function Computation for Dynamics with State Resets.}
 \label{numerical_algo}
\end{algorithm}

The numerical algorithm is implemented by means of the computational methods provided in \cite{mitchell2004toolbox}.
It is important to stress the remarkable computational similarity of this new algorithm to its non-reset counterpart in \eqref{eq:HJIVI}.
Indeed, the only computational overhead is introduced by the step in \eqref{constraint_in_algo}.
As a result, as will be demonstrated in the following section, our method can compute the BRT at essentially no additional cost compared to the case where there are no state resets.
\begin{remark} Note that Theorem \ref{theorem1} and the numerical method in Algorithm \ref{numerical_algo} are not specific to the RoA computation problem, which is the main focus of this paper. In general, these can be applied to any kind of target set $\targetset$ that can be expressed as a zero-sublevel set of a Lipschitz continuous function $l(x)$ and has a non-empty volume.
\end{remark}

\section{Case Studies} \label{sec:examples}
In this section, we will demonstrate how our extended HJ reachability framework can be used to compute RoA for various hybrid systems, including a simple biped walker. 

\subsection{Running example: Teleporting Dubins car} \label{sec:dubins_car:results}
We first apply our method to the running example. 
We use the following parameters for our simulations: $v\!=\!\SI{1}{\meter\per\second}$, $R\!=\!\SI{2}{\meter}$, $\bar{\omega}\!=\!\SI{1}{\radian\per\second}$, $\alpha\!=\!0.5$, which induces a contracting reset map, and $\epsilon=0.2$.

To confirm that our new method is able to capture states that exploit discrete jumps to reach the limit cycle, we compare our method to a simple baseline --  we compute the set of states that are stabilizable to the limit cycle only by following the continuous part of the dynamics. Thus, for this baseline, we ``freeze'' the dynamics on the switching surface, i.e. set $f(x, u)\equiv0$ whenever $x \in \switchsurface$.

\begin{figure}
\centering
\includegraphics[width=\columnwidth]{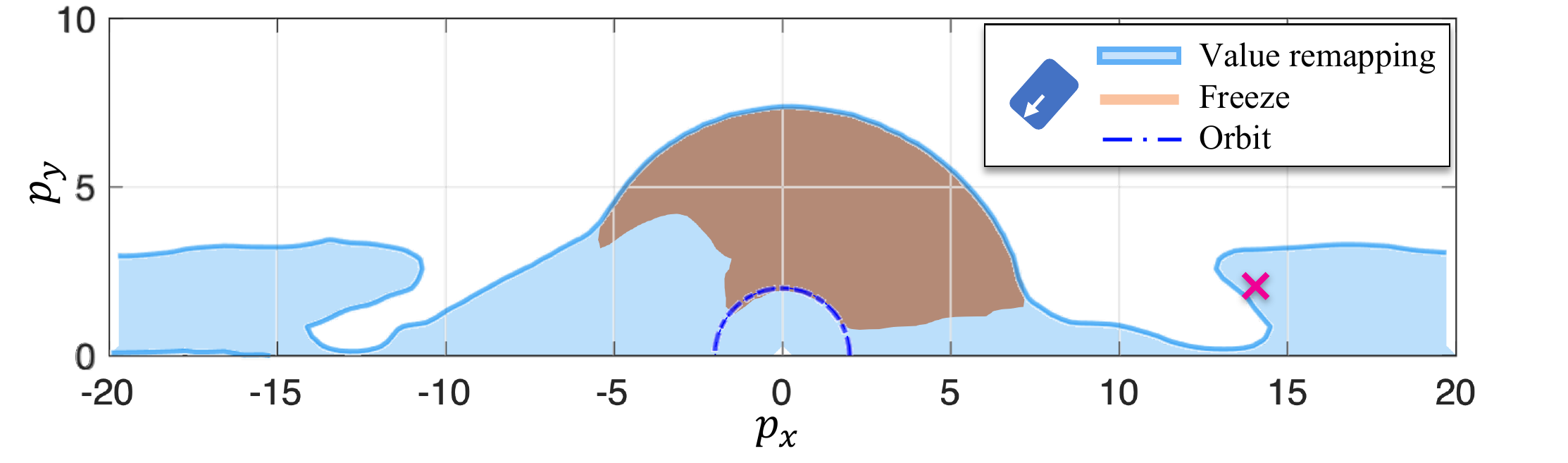}
\vspace{-1em}
\caption{$p_{x}$-$p_{y}$ slices of the Backward Reachable Tubes for the teleporting Dubins car at $\theta=-3\pi/4$, computed by value remapping (blue), and by freezing the dynamics (brown) on the switching surface ($T\!=\!\SI{6.3}{\second}$, $\alpha=0.5$). The gap between the two sets captures the states that can reach the target set within $\SI{6.3}{\second}$ only by using the ``teleports'' (reset map). We evaluate an initial state included in the BRT (cross) in Fig.\ref{fig:dubins_trajs}.
}
\label{fig:dubins_remapping_vs_freeze}
\vspace{-1.5em}
\end{figure}

\begin{figure}
\centering
\includegraphics[width=\columnwidth]{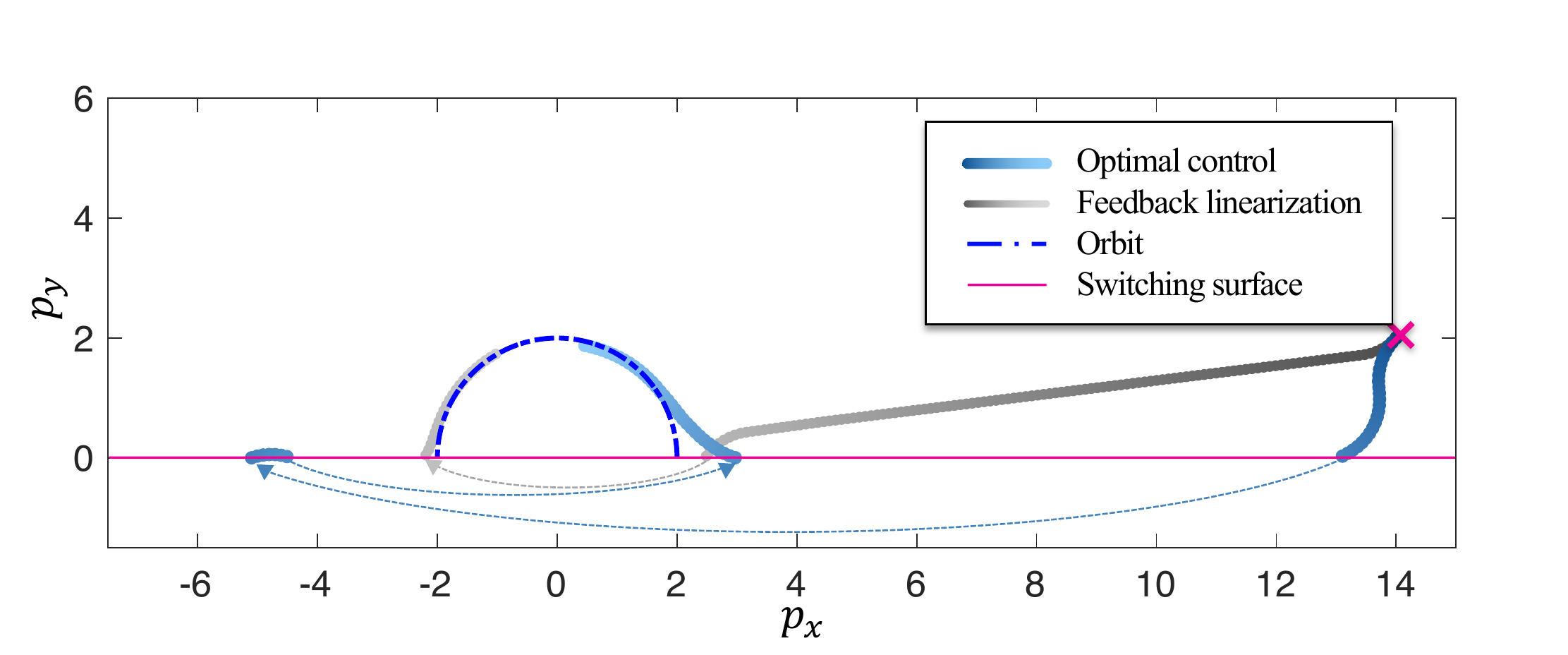}
\vspace{-1.5em}
\caption{The optimal trajectory simulated for $\SI{6.3}{\second}$ which reaches the target set after $\SI{4.3}{\second}$ by exploiting two discrete transitions. For comparison, the feedback-linearization-based baseline controller is simulated for $\SI{14}{\second}$, which reaches the target set after $\SI{11.9}{\second}$.
}
\label{fig:dubins_trajs}
\vspace{-2.0em}
\end{figure}

Fig.~\ref{fig:dubins_remapping_vs_freeze} visualizes the BRTs computed for $T\!=\!\SI{6.3}{\second}$ from our method (blue), and under the frozen dynamics (brown). It shows the 2D slices of the BRTs at a fixed heading angle $\theta=-\SI{3\pi/4}{\radian}$.  Note that the BRT computed by freezing the dynamics is strictly contained in the BRT computed by our method. The gap between the two sets uncovers the states that can reach the target set within $\SI{6.3}{\second}$ only via leveraging discrete jumps. Fig.~\ref{fig:dubins_trajs} displays the resulting optimal trajectory (blue) from a particular initial state that is contained in this gap (marked with the red cross). It exploits two state jumps to quickly reach the target set after $\SI{4.3}{\second}$; at every discrete jump, the distance to the orbit reduces significantly. In contrast, the feedback linearization-based baseline-controller (grey), although doing its best to get closer to the orbit by heading towards the radial direction, is able to reach the target set only after $\SI{11.9}{\second}$. Furthermore, the discrete jump near the limit cycle is not ``intended'' by the baseline controller, since the control law is not aware of the reset map. 
In contrast, the proposed method is \textit{explicitly} able to reason that since the reset map is a contraction in this case, it can be exploited to reduce the distance to the orbit and stabilize to the limit cycle faster.

\begin{figure}
\centering
\includegraphics[width=\columnwidth]{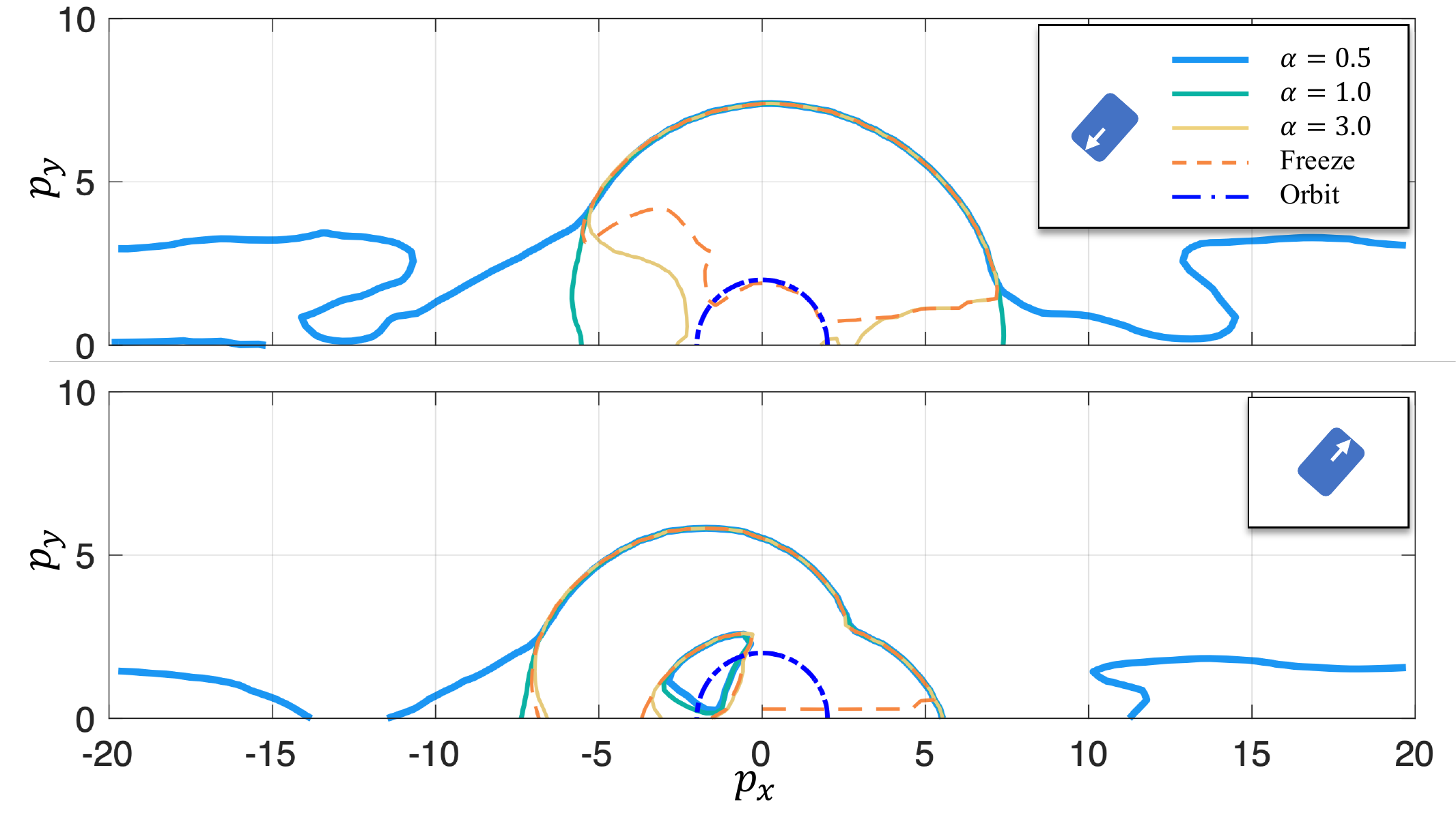}
\vspace{-.5em}
\caption{$p_{x}$-$p_{y}$ slices of the Backward Reachable Tubes at $\theta=-3\pi/4$ (top) and $\theta=\pi/4$ (bottom), with various values of the reset map nonlinearity parameter $\alpha$ under the value remapping, and under dynamics freezing, applied to the switching surface ($T\!=\!\SI{6.3}{\second}$). Note that the reset map is contracting when $\alpha<1$, and expanding when $\alpha>1$.
}
\label{fig:dubins_various_alphas}
\vspace{-.5em}
\end{figure}

In Fig.~\ref{fig:dubins_remapping_vs_freeze}, we can observe that many states adjacent to the switching surface are captured as the stabilizable region to the limit cycle. This is because $\alpha=0.5$ induces the contracting property of the reset map, which is beneficial for the stabilization. Fig.~\ref{fig:dubins_various_alphas} visualizes BRTs computed for the same $T$ under three different values of $\alpha$. Note that $\alpha=3.0$ is unfavorable for the stabilization as it is a divergent reset map; as a result, most of the region adjacent to the switching surface is not captured in the BRT. 
In fact, with even larger $\alpha$, the BRT eventually shrinks to the BRT computed by freezing the dynamics on the switching surface, i.e., the reset map is so divergent that it is no longer possible to stabilize the system after hitting the switching surface.

\textit{Computation details:} The grid used for the computation is constructed in the polar coordinate system of the dynamics ($r=\|(p_x, p_y)\|$, $\alpha=\tan^{-1}(p_y / p_x)$, $\theta$); it is a $81\!\times\!41\!\times\!81$ grid over the state space of $[0, 16]\!\times\![0, \pi]\!\times\![-\pi,\pi]$. It takes under an hour to compute the BRT of $T\!=\!\SI{6.3}{\second}$ on a computer with a 2.6GHz 6‑core 9th‑generation Intel Core i7 processor CPU.

\subsection{Rimless wheel system}
\begin{figure}
  \centering
  \includegraphics[width=0.95\columnwidth]{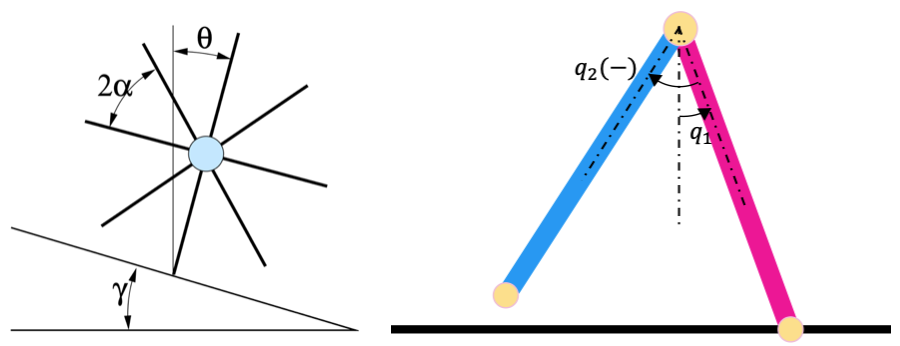}
  \caption{\footnotesize{(Left) Rimless wheel system. (Right) Compass-gait walker.}}
  \vspace{-0.75em}
  \label{fig.system_config}
\end{figure}
We next apply the proposed approach to the rimless wheel system. 
The rimless wheel is a popular model of a passive-dynamic walker.
One of the reasons for its popularity is that RoA can be computed analytically for the rimless wheel \cite{coleman1998stability}, and hence it can be used to analyze the efficacy of a RoA computation algorithm.

The system consists of a central mass with several spokes extending radially outward (see Fig. \ref{fig.system_config}); it has no control input.
At any given time, one of the spokes is pinned at the ground, and the system follows the dynamics of a simple pendulum, $\dyn(\theta, \dot{\theta})  = [\dot{\theta}, \sin(\theta)]^T,$
where $\state := (\theta, \dot{\theta})$ is the state of the system.
When an unpinned spoke contacts the ground, an inelastic collision takes place and there is an instantaneous change in $\dot{\theta}$. Upon collision, the pinned spoke leaves the ground and the new spoke becomes the pinned one.
The switching surface and the reset map are given as:
\vspace{-1em}

\small
\begin{equation*}
\switchsurface = \{\state \; | \theta = \alpha + \gamma \},\quad \poststate = \resetmap(\prestate^{-}) = \left[\begin{array}{c}
2\gamma - \theta \\
\cos(2\alpha)\dot{\theta}
\end{array}\right].
\end{equation*}
\normalsize
\begin{figure}
  \centering
  \vspace{0em}
  \includegraphics[width=\columnwidth]{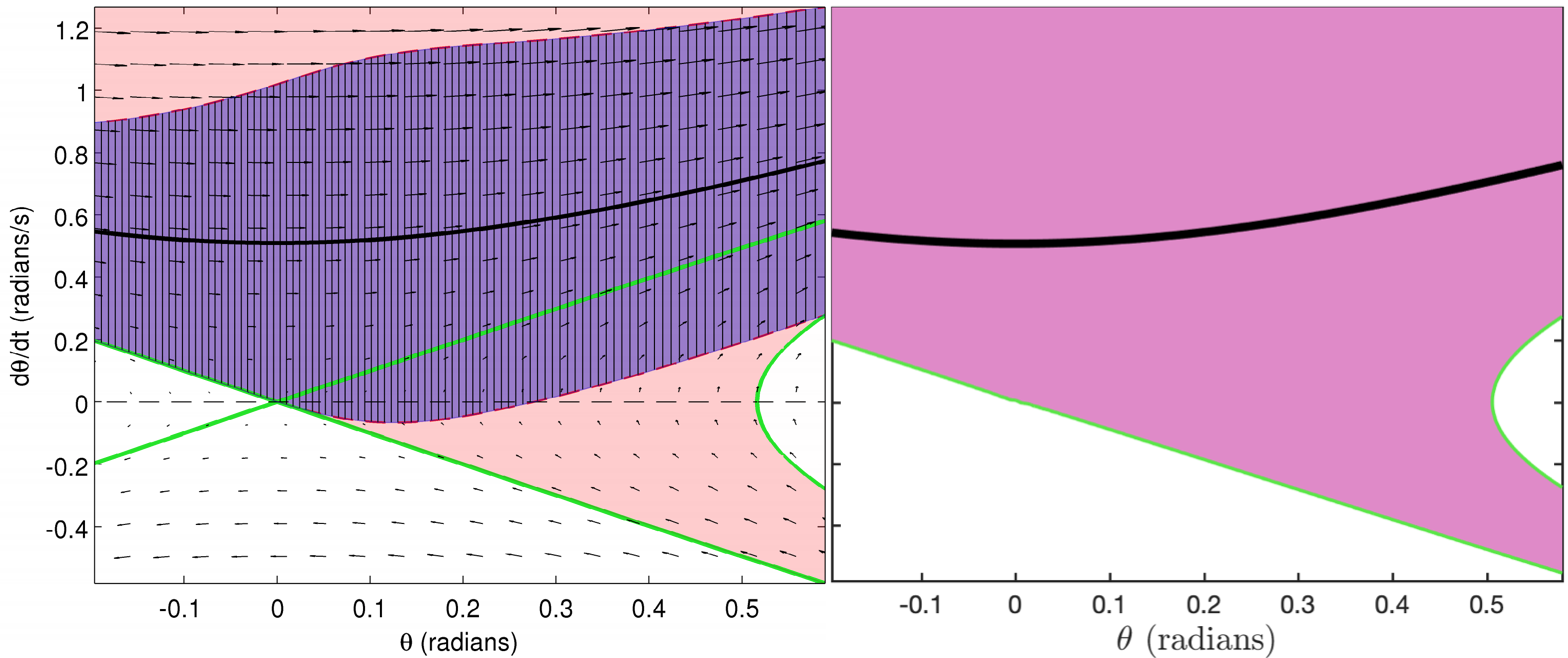}
  \caption{\footnotesize{Regions of Attraction (RoA) for the rimless wheel limit cycle (black). (Left) The true RoA is shown in light pink. RoA obtained using SOS programming is shown in purple. The figure is reproduced from \cite{manchester2011regions}. (Right) RoA computed using the proposed approach. The proposes approach is able to recover the entire RoA.}
  }
  \vspace{-2.25em}
  \label{fig.rimless.results}
\end{figure}

\vspace{-0.5em}
For a sufficiently inclined slope, the rimless wheel exhibits several stable hybrid limit cycles that are given as 

\vspace{-0.5em}
\small
\begin{equation*}
\limitcycle(\cdot) = \{\state \; | \cos(\theta) + \frac{1}{2}\dot{\theta}^2 = E \},
\end{equation*}
\normalsize
\vspace{-0.5em}

\noindent where $E$ is the total initial energy in the system.
One such limit cycle is shown in black in Fig.~\ref{fig.rimless.results} (left).
The figure is reproduced from \cite{manchester2011regions}.
The right boundary of the plot is the switching surface--whenever the system state reaches the right boundary, it resets back to the left boundary as per the reset map.
The light pink region is the exact RoA of the system \cite{coleman1998stability}.
The purple region is the RoA obtained using a time-varying Lyapunov function, which is synthesized using SOS programming \cite{manchester2011regions}.
As evident from the figure, using the SOS programming, we are able to recover only a subset of the RoA. In Fig.~\ref{fig.rimless.results} (right), we demonstrate the RoA obtained using our method (dark pink). The proposed approach is able to recover the entire RoA of the system.

\textit{Computation details:} We use a grid of size $201\times201$ over the state space $[-0.2,0.6]\times[-0.6, 1.3]$. 
It takes within a few minutes to compute the BRT on a standard laptop.

\subsection{Compass-gait walker}

We next consider a compass-gait walker, which consists of two links with an actuated joint between them. Similar to the rimless wheel example, we consider a \emph{pinned} model of the robot, with the configuration variable $q \coloneqq [q_1, q_2]^T$ as illustrated in Fig. \ref{fig.system_config}, 
We consider a hybrid model of walking, with alternating phases of a continuous-time single-support phase followed by an instantaneous inelastic impact of the swing leg with the ground. The dynamics in single-support can be represented by the standard manipulator equation, and the reset map $\resetmap$ is derived from an inelastic impact model \cite{westervelt2018feedback}. By defining the state as $\state \coloneqq [q, \dot{q}]^T$, the continuous dynamics can be expressed as control-affine: $\dot{x}=f(x)+g(x)u$. The switching surface is defined as the set of states where the swing foot strikes the ground with a negative velocity and the stance leg angle crosses a predefined threshold $\bar{q}_1$, 
\begin{align}
    \switchsurface := \lbrace \state ~|~ q_1 \leq \bar{q}_1, 2q_1 + q_2 = 0, 2\dot{q}_1 + \dot{q}_2 < 0 \rbrace.
\end{align}

\noindent

\begin{figure}\centering
\includegraphics[width=\columnwidth]{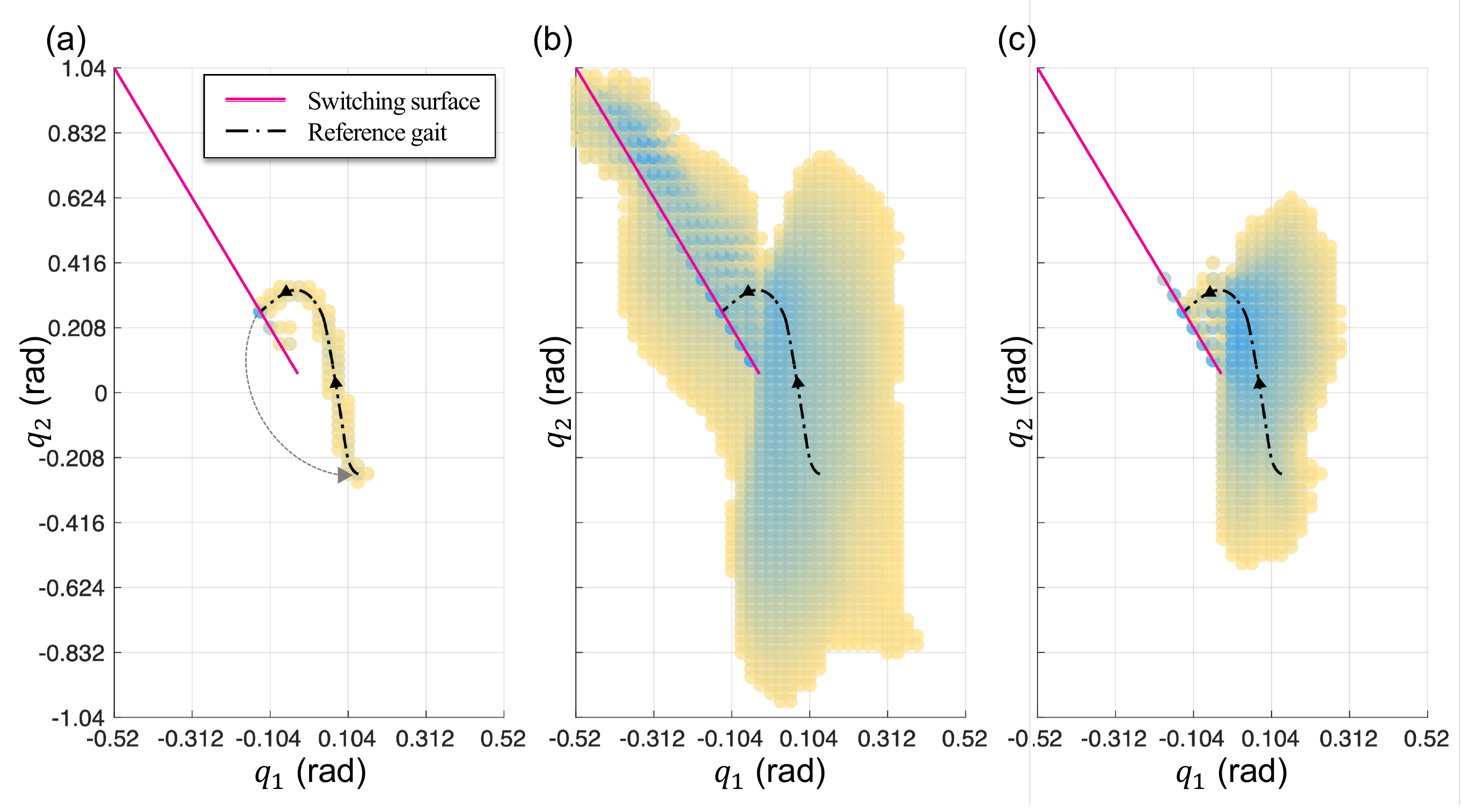}
\vspace{-1em}
\caption{Backward Reachable Tubes (BRTs) for the compass-gait walker visualized by their projection on the $q_1$--$q_2$ space. ($T\!=\!\SI{1.5}{\second}$ for all three cases) (a) BRT for the closed-loop dynamics under the CLF-QP. (b) BRT with the control bound $u\in[-4, 4]$, without the disturbance. (c) BRT with the control bound $u\in[-4, 4]$, and the disturbance bound $d\in[-0.3, 0.75]$. The color indicates the thickness of the BRT in the 4D space; it gets thicker as yellow$\rightarrow$blue.
}
\label{fig:compass_comparison_brts}
\vspace{-1.0em}
\end{figure}

The target hybrid limit cycle we wish to stabilize to is designed as a reference walking gait \cite{grizzle2001asymptotically,westervelt2003hybrid,hereid2017frost}, which is tracked by a baseline controller designed based on feedback linearization and CLF-QP \cite{ames2014rapidly}. The configuration of the swing leg on the gait, $q_{2,r}$, can be parametrized as a polynomial function of the stance leg angle $q_1$, and we define the output as $y:=q_{2}-q_{2,r}$ for feedback linearization. Each walking step takes $\SI{0.73}{\second}$, where $q_1$ starts with $q_{1,d}=0.13$ and ends at $-q_{1, d}$. The phase plot of the reference gait in $q_1$-$q_2$ space is visualized in Fig. \ref{fig:compass_comparison_brts}(a) black dashed line.

\begin{figure}\centering
\includegraphics[width=\columnwidth]{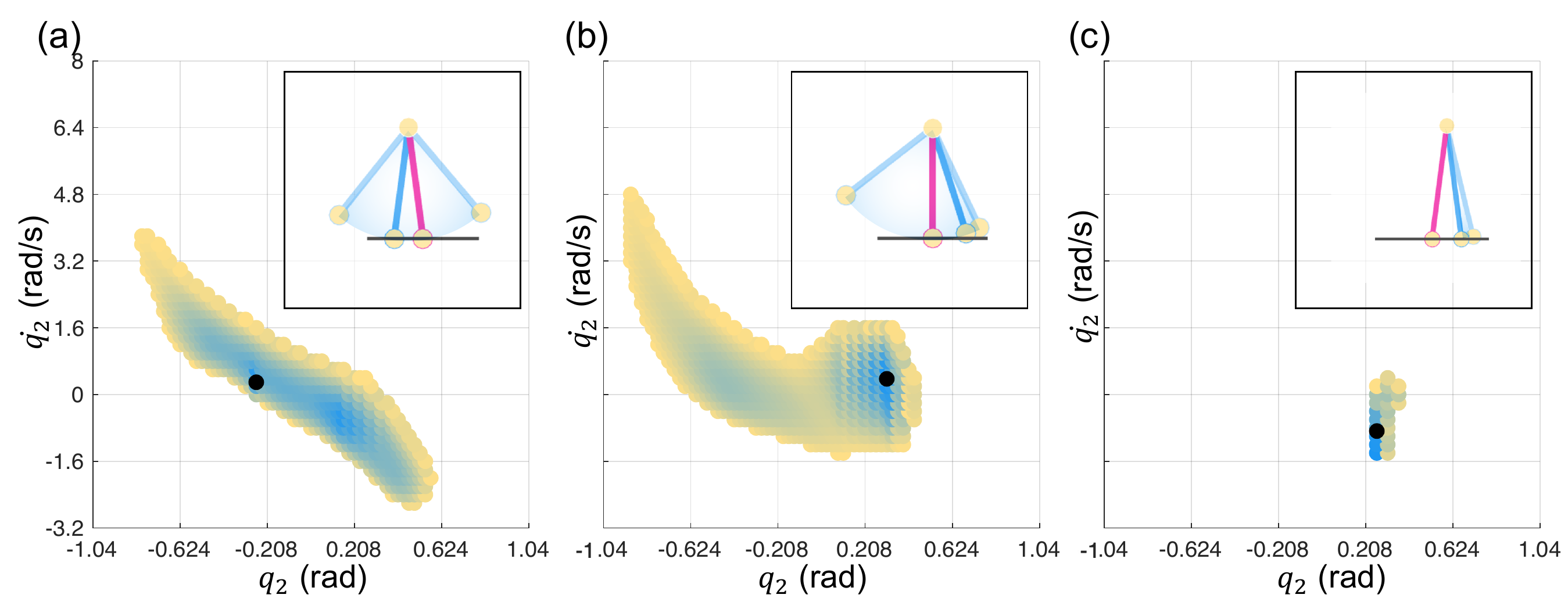}
\vspace{-1.5em}
\caption{$q_2$--$\dot{q_2}$ slices of the BRT corresponding to Figure \ref{fig:compass_comparison_brts}-(a), along the hybrid periodic orbit (reference gait). (a) Initial phase: $[q_1, \dot{q_1}]=[0.13, -0.67]$. (b) Intermediate phase: $[q_1, \dot{q_1}] =[0, -0.36]$. (c) Terminal phase (imminent of reset): $[q_1, \dot{q_1}] =[-0.13, -0.91]$. The black dots indicate the corresponding $[q_2, \dot{q_2}]$ values on the orbit, which are always contained in the BRT. The color indicates the value of $V(x, T)$; it is yellow when $V(x, T)=0$ and gets blue as the value get smaller. The diagrams on the top right corners visualize the swing leg (blue) configuration range that is contained in the BRT, together with the stance leg (magenta) for each phase.
}
\label{fig:compass_brt_config}
\vspace{-2.0em}
\end{figure}

In Fig. \ref{fig:compass_comparison_brts}, the BRTs computed from our method for $T\!=\!\SI{1.5}{\second}$ is visualized as the projection of the 4D set to the $q_1$-$q_2$ space (colored region). Note that the chosen time horizon encompasses more than two nominal walking steps. First, we apply our method to the closed loop dynamics of the baseline controller (Fig. \ref{fig:compass_comparison_brts}(a)). Although it is designed to be locally stabilizing, the verified RoA comprises only a small neighborhood of the reference gait. By contrast, Fig. \ref{fig:compass_comparison_brts}(b) shows the verified stabilizable region for the gait by allowing full control authority defined by the control bound $u\in[-4,4]\SI{}{\newton}$. The bound is selected such that on the limit cycle, the CLF-QP barely saturates this bound, so that we do not give additional advantage to the optimal controller. However, by exploiting full control capacity and the reset map, we are able to verify that the actual stabilizable region is much bigger than what can be achieved by the CLF-QP controller. Fig.~\ref{fig:compass_brt_config} shows the $q_2$--$\dot{q_2}$ slices of the BRT along the reference gait. Note that the gait itself is actually contained in the BRT, and that the BRT gets smaller as it gets closer to the end of the step (Fig.\ref{fig:compass_brt_config}(c)), because of the imminent reset.

\begin{figure}\centering
\includegraphics[width=\columnwidth]{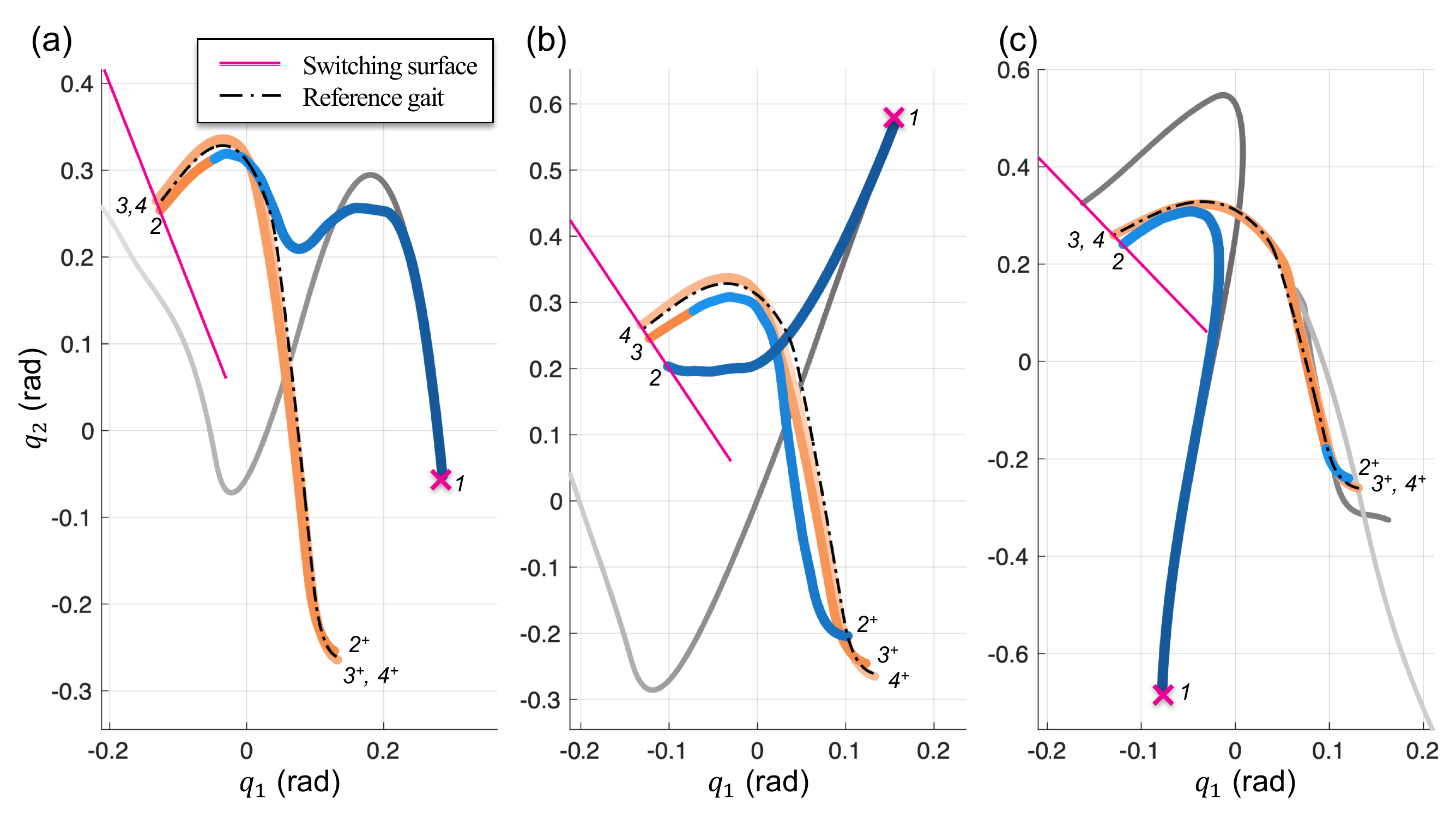}
\vspace{-1.5em}
\caption{Phase plots of the reachability-based optimal trajectories (blue$\rightarrow$orange) and the CLF-QP trajectories (grey) in $q_1$--$q_2$ space for various initial states from which the CLF-QP fails to stabilize to the gait. The trajectories are evaluated for 4 walking steps. Each start of the step is indicated by the index of the step in italic type. The evolution of time is indicated by the color fading out. The optimal trajectory switches from HJ optimal control to CLF-QP once it reaches the target set (indicated by the transition from blue to orange curve). 
}
\label{fig:compass_opt_trajs}
\vspace{-1.0em}
\end{figure}

In Fig. \ref{fig:compass_opt_trajs}, we show some trajectories that start at initial states which are contained in the verified BRT. These initial states have a large deviation from the reference gait, corresponding to a large perturbation. We start by applying the reachability-based optimal control (blue), and then switch to the CLF-QP controller (orange) when we reach the target set. In Fig. \ref{fig:compass_opt_trajs}(b,c), the optimal controller is exploiting the reset map to converge to the limit cycle faster. Physically, this means that the robot steps on the ground first, and use this ground force to converge to the nominal gait at the next walking step. For comparison, the trajectories from the same initial states where the CLF-QP is used from the very beginning are visualized together (grey), which all fail to recover from the ground impact and cannot stabilize to the limit cycle. In particular, in Fig.~\ref{fig:compass_opt_trajs}(c), the CLF-QP fails to overcome a large momentum, resulting in the robot to fall backwards.

\begin{figure}\centering
\includegraphics[width=\columnwidth]{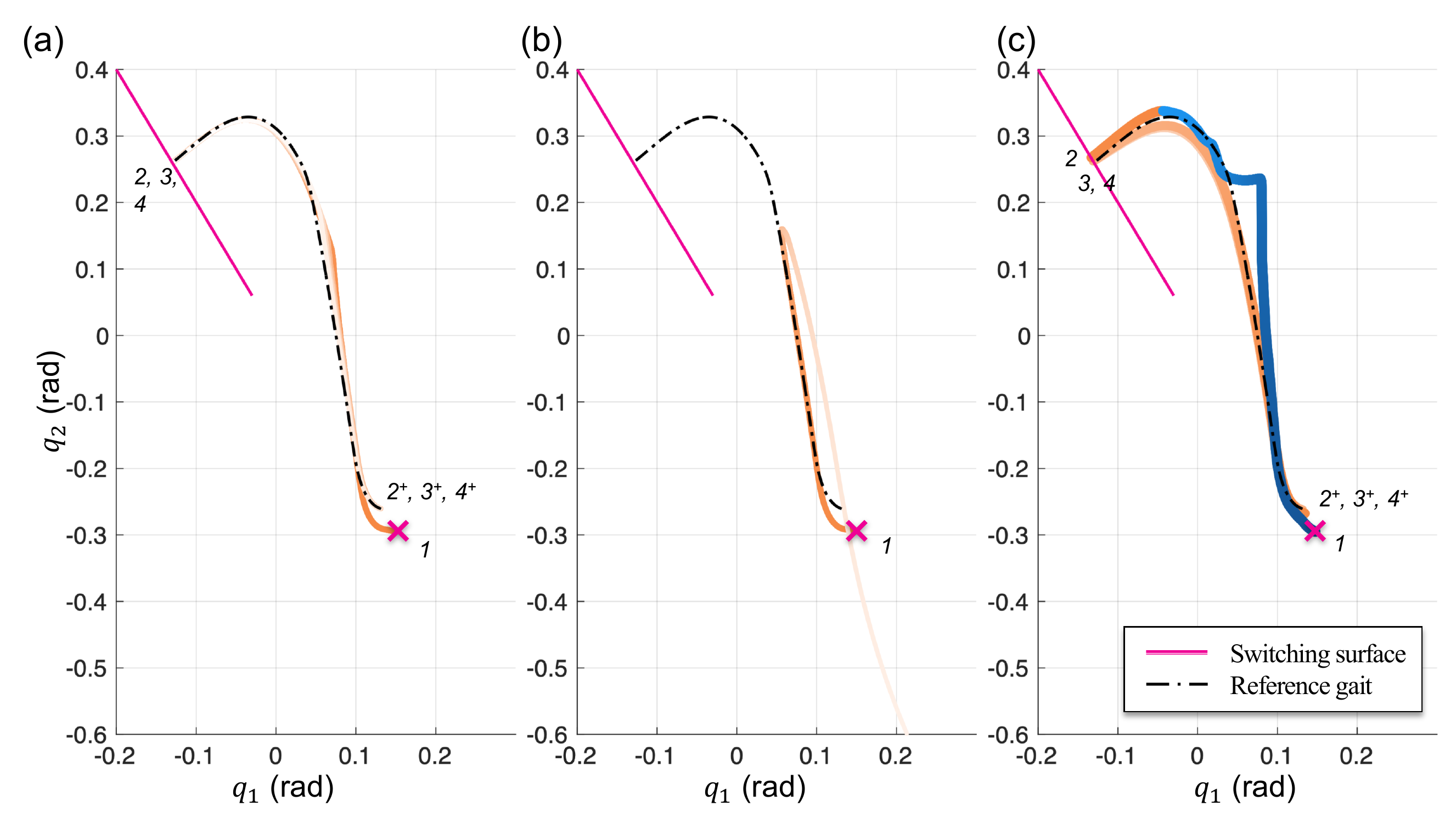}
\vspace{-1.5em}
\caption{Phase plots of the trajectories in $q_1$--$q_2$ space evaluated for 4 walking steps for a same initial state: (a) CLF-QP under no model-plant mismatch. (b) CLF-QP under model-plant mismatch, which fails to stabilize to the orbit. (c) Optimal control, which is able to reach the orbit under the same model-plant mismatch. Each start of the step is indicated by the index of the step in italic type. The evolution of time is indicated by the color fading out.
}
\label{fig:compass_compare_disturbance}
\vspace{-2.0em}
\end{figure}

Finally, we introduce disturbance to the robot's dynamics: an unmodeled repulsive or stiction torque applied to the joint between the leg. This can be expressed as $\dot{x}=f(x)+g(x)u+\dot{q}_2 g(x) d$. 
There could be various source of uncertainty for this disturbance, for instance, unmodeled motor dynamics or joint friction. The bound of the disturbance is $d\in[-0.75, 0.3]$; under this bound, on the limit cycle, the CLF-QP is still able to robustly stabilize the gait. The BRT computed under the robust optimal control setting (as explained in Sec. \ref{subsec:disturbance}) is visualized in Fig.~\ref{fig:compass_comparison_brts}.(c). Since now we only verify states that are \textit{robustly} stabilizable to the gait, the resulting BRT is smaller than the non-robust BRT in Fig. \ref{fig:compass_comparison_brts}.(b). However, it still shows that by applying the robust optimal controller, it is able to robustly stabilize a larger set of states than the CLF-QP. For instance, the CLF-QP fails to stabilize a state slightly perturbed from the gait under the presence of disturbance $d\!=\!-0.75$ (Fig.\ref{fig:compass_compare_disturbance}.(b)), because the CLF-QP cannot reason about the additional torque required to compensate the effect of the disturbance. 
In contrast, the robust stabilizing controller obtained using the proposed approach is simultaneously able to counter the disturbances and stabilize to the reference gait (Fig.\ref{fig:compass_compare_disturbance}.(c)). 

\textit{Computation details:} We use a grid of size $41\times81\times81\times81$ over the state space of $[-0.52,0.52]\times[-1.04, 1.04]\times[-4, 4]\times[-8, 8]$, taking roughly 36 hours to compute the BRT of $\SI{1.5}{\second}$ for this demonstration using a 3.3GHz 12-core Intel Core i7-5820K CPU.

\section{Conclusion and Future Work} \label{sec:conclusion}
We present a reachability-based approach to compute regions-of-attraction for dynamical systems with state resets. 
We apply our approach to several underactuated walking robots and demonstrate that the proposed approach can synthesize a bigger RoA compared to the state-of-the-art approaches and also provides a stabilizing controller.
However, the proposed approach has several limitations that we will like to address in future.
First, the reachability computation scales exponentially with the number of states, limiting its direct use to relatively low-dimensional systems.
Recently, there have been some promising results that use Neural PDE solvers for computing reachable set for high-dimensional systems \cite{bansal2020deepreach}. 
We will explore using these methods to extend the proposed approach to more complex walking robots.
For providing stability guarantees, it would be interesting to leverage recent sampling-based methods \cite{boffi2020learning} that provide probabilistic guarantees on the learned value functions. Second, it will be interesting to extend the proposed approach to other types of reachability problems, such as reaching a target set while avoiding some undesirable states, e.g., obstacles.
Applying the proposed approach to real-world robotic systems will be another exciting future direction.

\printbibliography

\newpage
\section*{Appendix A} \label{sec:appendix1}
\subsection{Proof of Theorem \ref{theorem1}}
We will first prove the result for a state that is not on the switching surface. 
Let $\prestate(\tvar) \notin \switchsurface$. 
The result for this case follows from the proof of standard HJI-VI.
Here we present an informal proof based on the Taylor series expansion of the value function; a more rigorous proof of HJI-VI can be found in \cite{Mitchell05}.

Consider a small $\delta > 0$ such that there is no state reset in the time interval $[\tvar, \tvar+\delta]$.
The dynamic programming principle for the cost function in \eqref{eq:costfunctional} implies that

\small
\vspace{-1.2em}
\begin{align} \label{eq:proof_help1}
\vfunc(\state(\tvar), \tvar) & = \sup_{\ctrl \in \cset} \inf_{\dstb \in \dset} \min\{\targetfunc(\state(\tvar)), \vfunc(\state(\tvar+\delta), \tvar+\delta)\}, \nonumber \\
& =  \min\{\targetfunc(\state(\tvar)), \sup_{\ctrl \in \cset} \inf_{\dstb \in \dset} \vfunc(\state(\tvar+\delta), \tvar+\delta))\},
\end{align}
\normalsize
On the other hand, the Taylor expansion of $\vfunc(\state(\tvar+\delta), \tvar+\delta)$ can be written as:

\small
\vspace{-1.2em}
\begin{equation*}
\vfunc(\state(\tvar+\delta), \tvar+\delta) = \vfunc(\state(\tvar), \tvar) + D_\tvar\vfunc(\state(\tvar), \tvar)\delta + \nabla\vfunc(\state(\tvar), \tvar)\delta\state + \text{h.o.t},
\end{equation*}
\normalsize
where $\delta\state$ is change in the state and can be approximated as $\dyn(\state, \ctrl, \dstb)\delta$.
%
%
Ignoring the higher order terms and plugging the Taylor approximation in \eqref{eq:proof_help1}, we have

\small
\vspace{-1.2em}
\begin{align*}
\vfunc(\state(\tvar), \tvar) \approx \min\{\targetfunc(\state(\tvar)), & \vfunc(\state(\tvar), \tvar) + D_\tvar\vfunc(\state(\tvar), \tvar)\delta \\
& + \sup_{\ctrl \in \cset} \inf_{\dstb \in \dset} \nabla\vfunc(\state(\tvar), \tvar)\cdot\dyn(\state, \ctrl, \dstb)\delta \}, 
\end{align*}
\normalsize
where we have separated terms that do not depend on $\ctrl$ and $\dstb$.
Subtracting $\vfunc(\state(\tvar), \tvar)$ on both sides, we get

\small
\vspace{-1.2em}
\begin{align*}
\min\{&\targetfunc(\state(\tvar)) - \vfunc(\state(\tvar), \tvar),\\
&\delta \left[D_\tvar\vfunc(\state(\tvar), \tvar) + \sup_{\ctrl \in \cset} \inf_{\dstb \in \dset} \nabla\vfunc(\state(\tvar), \tvar)\cdot\dyn(\state, \ctrl, \dstb)\right] \} \approx 0, 
\end{align*}
\normalsize
Since the above equation holds for all $\delta > 0$, we must have

\small
\vspace{-1.2em}
\begin{align*}
\min\{&\targetfunc(\state(\tvar)) - \vfunc(\state(\tvar), \tvar),\\
&\left[D_\tvar\vfunc(\state(\tvar), \tvar) + \sup_{\ctrl \in \cset} \inf_{\dstb \in \dset} \nabla\vfunc(\state(\tvar), \tvar)\cdot\dyn(\state, \ctrl, \dstb)\right] \} \approx 0, 
\end{align*}
\normalsize
which results into the desired HJI VI.

We now turn our attention to a state $\prestate(\tvar) \in \switchsurface$.
Once again, the dynamic programming principle for the cost function in \eqref{eq:costfunctional} implies that

\small
\vspace{-1.2em}
\begin{align*}
\vfunc(\state(\tvar), \tvar) & = \sup_{\ctrl \in \cset} \inf_{\dstb \in \dset} \min\{\targetfunc(\state(\tvar)), \vfunc(\state(\tvar+\delta), \tvar+\delta)\}, \end{align*}
\normalsize
Since $\prestate(\tvar) \in \switchsurface$, the system state will instantaneously change to $\poststate(\tvar)$ under any control and disturbance action.
Thus, we have

\small
\vspace{-1.2em}
\begin{align*}
\vfunc(\state(\tvar), \tvar) & = \sup_{\ctrl \in \cset} \inf_{\dstb \in \dset} \min\{\targetfunc(\poststate(\tvar)), \vfunc(\poststate(\tvar+\delta), \tvar+\delta)\}, \\
& = \vfunc(\poststate(\tvar), \tvar), 
\end{align*}
\normalsize
where the second equality follows from the definition of value function for $\poststate(\tvar)$.

\end{document}